\documentclass[11pt]{article}

\usepackage[preprint]{acl}

\usepackage{times}
\usepackage{latexsym}

\usepackage[T1]{fontenc}

\usepackage[utf8]{inputenc}

\usepackage{microtype}

\usepackage{inconsolata}

\usepackage{graphicx}

\usepackage{amsmath}
\usepackage{bbm}
\usepackage{enumitem}
\usepackage{booktabs}
\usepackage{makecell}
\usepackage{multirow}
\usepackage{tabularx}
\usepackage{pifont}
\usepackage{xcolor}
\usepackage{tcolorbox}
\usepackage{listings}
\tcbuselibrary{listings, breakable}
\usepackage{caption}
\usepackage{colortbl}
\usepackage{amssymb}

\newcommand{\cmark}{\textcolor{green!60!black}{\ding{51}}}
\newcommand{\xmark}{\textcolor{red!80!black}{\ding{55}}}

\setlist[itemize]{itemsep=2pt, parsep=0pt, topsep=0pt, partopsep=0pt}

\title{LifePlanner: Evaluating LLM Agents for Geo-spatial Planning with Social Media Data
}

\author{
  \textbf{Zhen Dong\textsuperscript{1,*}},
  \textbf{Yuning Peng\textsuperscript{1,*}},
  \textbf{Yutao Shi\textsuperscript{1}},
  \textbf{Lei Zhong\textsuperscript{1}},
\\
  \textbf{Yongsen Mao\textsuperscript{2}},
  \textbf{Yuan Liu\textsuperscript{2}},
  \textbf{Haiping Wang\textsuperscript{1,2}}
\\
\\
  \textsuperscript{1}Wuhan University
  \qquad
  \textsuperscript{2}Hong Kong University of Science and Technology
\\
  \small{\texttt{\{dongzhenwhu,yuningpeng,yutaoshi,leizhong\}@whu.edu.cn}}
\\
  \small{\texttt{yongsen.mao@connect.ust.hk}, \texttt{\{yuanly,hpwang\}@ust.hk}}
\\
  \small{\textsuperscript{*}Equal contribution. \quad \textbf{Correspondence:} \href{mailto:hpwang@ust.hk}{hpwang@ust.hk}}
}

\begin{document}
\maketitle
\begin{abstract}
Geo-spatial planning, like trip design, is a realistic testbed for LLM agents because it requires grounded tool use, noisy evidence retrieval, and multi-constraint reasoning. Most benchmarks, however, only provide clean geospatial data and tools, missing the open-ended social signals that people use in daily planning. We introduce LifePlanner, a benchmark that enriches map data with large-scale local social media posts and provides access through an MCP toolset. LifePlanner provides an evaluation suite spanning four task categories and three difficulty levels. Experiments show frontier LLMs perform well on simple retrieval but degrade sharply on complex planning, with the Pass Rate dropping to 40.2\%. Results show that failures mainly stem from incomplete evidence acquisition from such a large multimodal database, imprecise tool use, and weak constraint integration rather than model size or reasoning length, suggesting that future progress requires effective grounded planning instead of scaling alone.
\end{abstract}

\section{Introduction}

Large Language Models (LLMs) are evolving from passive text generators into agentic systems capable of tool use and long-horizon reasoning \citep{parisi2022talm, shen2023hugginggpt, patil2024gorilla}. Geo-spatial planning has therefore become an increasingly compelling task for evaluating such systems~\citep{MapEval, cheng2026travelbench, song2026mobilitybench}. 
Its core appeal lies in its comprehensive evaluation of three capabilities: long-context processing over multimodal and heterogeneous spatial evidence, tool use for grounded spatial analysis, and constrained reasoning for cost--preference trade-offs and spatially faithful planning.

We define geo-spatial planning as the process of selecting and ordering real-world places under geographic, temporal, semantic, and user-specific constraints using external spatial evidence and tools. Consider a user driving from an office to meet a friend who wants to stop at a venue where they can interact with short-legged dogs, visit a popular lakeside sunset spot, arrive while the venues are open, and minimize total driving time. Solving this request requires the agent to ground informal preferences in noisy social-media evidence, identify the corresponding places, verify temporal constraints, invoke routing tools, and jointly optimize the visiting order. Together, these steps characterize the coupled evidence-grounding and planning capabilities evaluated by LifePlanner.

\begin{table}[t]
\centering
\setlength{\tabcolsep}{2pt} 
\resizebox{\linewidth}{!}{
\begin{tabular}{l cc cc}
\toprule
\multirow{2}{*}{\raisebox{-0.6em}{\textbf{Work}}} & \multicolumn{2}{c}{\textbf{Env.}} & \multicolumn{2}{c}{\textbf{Eval.}} \\
\cmidrule(lr){2-3} \cmidrule(lr){4-5}
& \textit{\makecell{Tool\\Box}} & \textit{\makecell{Social\\Media}} & \textit{\makecell{Multi\\Task}} & \textit{\makecell{Multi\\Diff.}} \\
\midrule
\midrule
TravelPlanner \cite{xie2024TravelPlanner} & \cmark & \xmark & \xmark & \cmark \\
CityEval \cite{Feng2025CityGPT}           & \xmark & \xmark & \cmark & \xmark \\
USTBench \cite{lai2025ustbench}           & \xmark & \xmark & \cmark & \xmark \\
CityBench \cite{feng2025CityBench}        & \xmark & \xmark & \cmark & \xmark \\
MapEval \cite{MapEval}                    & \cmark & \xmark & \cmark & \xmark \\
TripScore \cite{qu2025tripscore}          & \xmark & \xmark & \xmark & \xmark \\
TP-RAG \cite{ni-etal-2025-tprag}          & \xmark & \xmark & \xmark & \xmark \\
TripTailor \cite{wang-etal-2025-triptailor}& \cmark & \xmark & \xmark & \cmark \\
ChinaTravel \cite{shao2026chinatravel}    & \cmark & \xmark & \xmark & \cmark \\
COMPASS \cite{qin2026compass}             & \cmark & \xmark & \xmark & \cmark \\
MobilityBench \cite{song2026mobilitybench}& \cmark & \xmark & \cmark & \xmark \\
DeepPlanning \cite{zhang2026deepplanning} & \cmark & \xmark & \xmark & \xmark \\
TravelBench \cite{cheng2026travelbench}   & \cmark & \xmark & \cmark & \xmark \\
\midrule
\textbf{LifePlanner (Ours)}               & \cmark & \cmark & \cmark & \cmark \\
\bottomrule
\end{tabular}
}
\vspace{-5pt}
\caption{Comparison with prior geo-spatial benchmarks. \textit{Multi Task} indicates the inclusion of varying geo-spatial task types, and \textit{Multi Diff.} denotes the presence of hierarchical task difficulty levels.}
\vspace{-15pt}
\label{tab:benchmark_comparison}
\end{table}

However, as summarized in Table~\ref{tab:benchmark_comparison}, existing geo-spatial planning benchmarks remain limited along two key dimensions. The first is the \textit{environment}, i.e., the database and tools available to LLMs. Most benchmarks are built on clean geospatial databases or manually curated resources \citep{shao2026chinatravel, ni-etal-2025-tprag, qu2025tripscore, MapEval, cheng2026travelbench, song2026mobilitybench, chen2025spatialllmm}. Some expose LLMs to map-based tools for information retrieval and analysis, yet they rarely incorporate \textit{open-ended social signals such as social-media posts, which are often central to how real users acquire timely, noisy, and locally grounded spatial knowledge}. The second is the \textit{evaluation protocol}. Existing tasks are often narrow in intent and weakly stratified in difficulty, making it hard to derive systematic trends and insights across planning tasks and complexity levels.

To address these limitations, we introduce {LifePlanner}, a scalable framework and benchmark for geo-spatial planning in realistic daily scenarios. At the \textit{environment} level, LifePlanner augments standard geospatial information with large-scale, region-specific social-media data and provides agents with tool access for grounded, up-to-date spatial analysis. At the \textit{evaluation} level, LifePlanner defines a multi-task benchmark spanning four daily planning categories, progressively moving from point-level to area-level planning: Place Perception, Nearby Discovery, Routing, and Trip Design. Each category is further organized into three structured difficulty levels: L0 requires a single search over the given database, L1 requires multiple retrievals for calculation, and L2 requires decision-making over retrieved evidence under additional user constraints. This design enables systematic assessment across both task diversity and planning complexity.

LifePlanner reveals three key limitations of current LLM agents in realistic urban planning. (1) Task difficulty exposes a sharp gap between simple retrieval and complex constrained planning. (2) Failures arise less from hallucination than from \textit{incomplete evidence acquisition and weak constraint integration}. (3) Model scaling and longer reasoning traces are insufficient; future agents need stronger training for effective tool use, targeted exploration over a large multimodal database, and faithful evidence-grounded planning.
To ensure reproducibility and robust measurement, we will open-source all data-collection tools and publicly release the processed, anonymized benchmark data.

\begin{figure*}[t!]
  \centering
  \includegraphics[width=\linewidth]{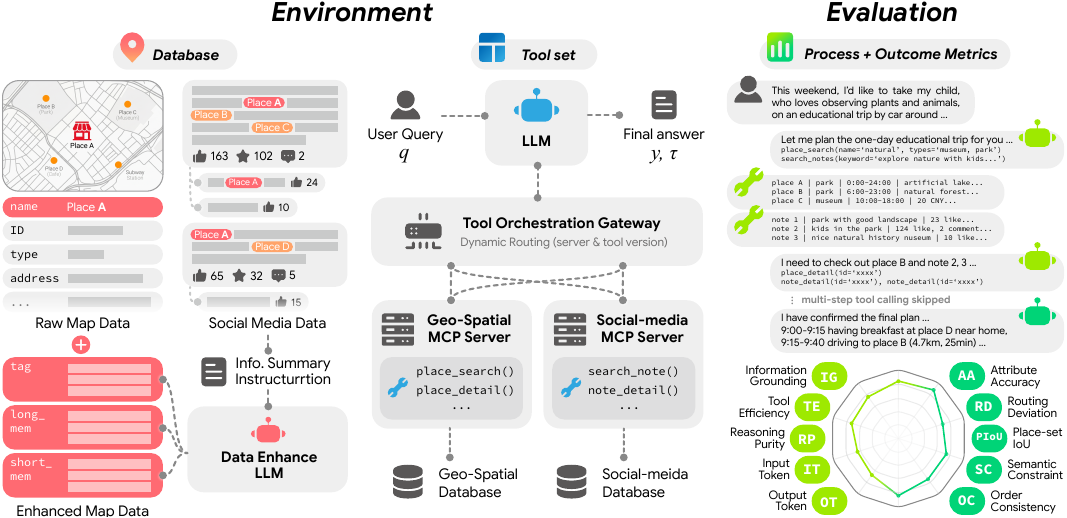}
  \caption{\textbf{LifePlanner Overview.} 
  LifePlanner consists of two main components: an environment and an evaluation protocol. The environment provides geospatial and social-media evidence through an MCP server, while the evaluation assesses both the tool-use process and final outputs of LLM agents.
  }
  \vspace{-10pt}
  \label{fig:teaser}
\end{figure*}

\section{Related Works}
\subsection{LLM Agents in City Scenario Planning}
Urban scenario planning severely challenges large language models (LLMs) due to diverse tasks and heterogeneous information. Previous studies have demonstrated that relying solely on parametric knowledge is fundamentally unreliable for complex geospatial reasoning \citep{wang2024isapicureworthathousandwords, yuan-etal-2025-llmap}. To address this, existing research predominantly pursues domain knowledge injection or external information augmentation. 

Under knowledge injection, prior work utilizes instruction-tuning with structured city-level data \citep{Feng2025CityGPT}, or incorporates multimodal inputs (e.g., street and satellite views) to broaden spatial capabilities \citep{Feng2025UrbanLLaVA,wang2025urbanr1,chu2026travelllama}. However, despite achieving a certain degree of generalization on specific spatial tasks, inherent domain discrepancies severely limit cross-city applicability, necessitating resource-intensive, city-specific fine-tuning.

Alternatively, external augmentation bypasses parametric limitations via Retrieval-Augmented Generation (RAG) and dynamic tool invocation. RAG frameworks integrate external data into the context, tailored for geo-spatial tasks via SQL-based filtering \citep{yu2025spatialrag} and document-based itinerary planning \citep{ni-etal-2025-tprag}. Extending beyond static retrieval, tool augmentation grants agents executable actions to dynamically acquire information \citep{xie2024TravelPlanner}. To simplify complex tool utilization, recent works decompose queries into sequential steps for specialized sub-agents \citep{zhe2025constraintawarerouterecommendation,hasan-etal-2026-mapagent}. Moving beyond pure geographic planning, the latest developments leverage user profiles for preference-aligned planning \citep{Liu2025LLM-TripPlanner,lan2025localgpt}. 

In this work, alongside user-side preference integration, we significantly expand the information capacity and density on the environment side. By designing a customized, comprehensive toolset, we empower the agent to not only query standard cartographic platforms but also dynamically acquire real-time insights from social media. This paradigm shift authentically simulates the inherent noise, richness, and complexity of planning tasks within real-world urban spaces.

\subsection{Evaluation of Geo-Spatial Reasoning}

To evaluate internalized capabilities of language models, CityEval \citet{Feng2025CityGPT} assesses text-centric spatial tasks, while USTBench \citep{lai2025ustbench} and CityBench \citep{feng2025CityBench} extend this to complex decision-making and multimodal data.
Conversely, external augmentation benchmarks evaluate agents using retrieved data or dynamic tools. TripScore and TP-RAG \citep{qu2025tripscore, ni-etal-2025-tprag} adopt a static context approach, providing necessary information upfront. Moving toward dynamic interaction, MapEval \citep{MapEval} and TravelPlanner \citep{xie2024TravelPlanner} require active data fetching via cartographic or domain-specific APIs. Recent works further assess interactive multi-turn preference inference \citep{qin2026compass, cheng2026travelbench}, multi-constraint optimization \citep{shao2026chinatravel, wang-etal-2025-triptailor, zhang2026deepplanning}, and fine-grained, real-world mobility planning \citep{song2026mobilitybench}.

Despite these advancements, existing benchmarks predominantly restrict agents to highly refined, structured data sources (e.g., cartographic databases or dedicated search APIs), which diverges sharply from actual human behavior. In reality, individuals invest significant effort in exploring noisy, unstructured data sources, such as social media, and deeply integrating this information into their daily planning. 
To bridge this gap, LifePlanner introduces socially-driven preferences (e.g., locating a "hidden gem" spot or limited-time events) that traditional structured platforms alone cannot satisfy. 
By compelling agents to actively navigate simulated social media and distill actionable information from high-noise environments, our benchmark pushes evaluation significantly closer to authentic real-world life planning.

\section{LifePlanner}
LifePlanner consists of two core components: a realistic \textit{environment} for LLM interaction and a systematic \textit{evaluation} protocol for measuring planning capability. The environment covers a database of approximately 10~km$^2$ of urban space, containing over 3,600 locations and about 200,000 social-media notes and comments, with tools for data retrieval and basic analysis. The evaluation protocol includes 667 synthetic queries expressing plausible daily planning needs and goals across four task categories and three complexity levels; for each, LLMs return a structured JSON output for multi-perspective scoring.

\begin{figure}
  \centering
  \includegraphics[width=\linewidth]{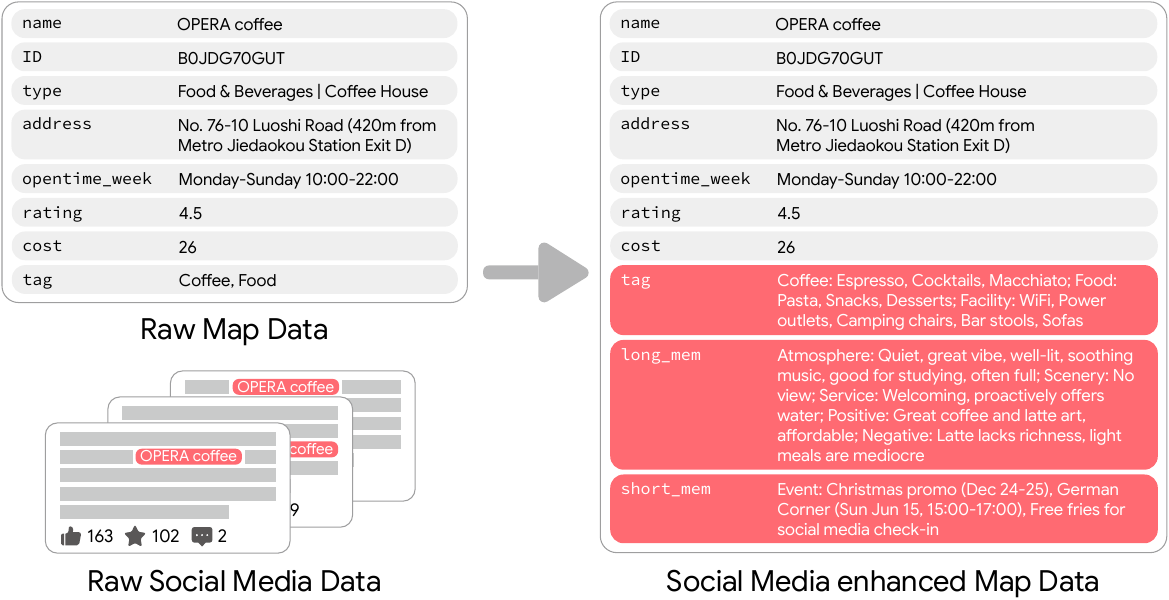}
  \caption{LifePlanner uses social media data to enrich map data and construct a rich, realistic environment. Note that evaluated LLMs cannot access the time-sensitive information (mem field) extracted from social media.}
  \label{fig:data_enhance}
\vspace{-10pt}
  
\end{figure}

\begin{figure*}[t!]
  \centering
  \includegraphics[width=\linewidth]{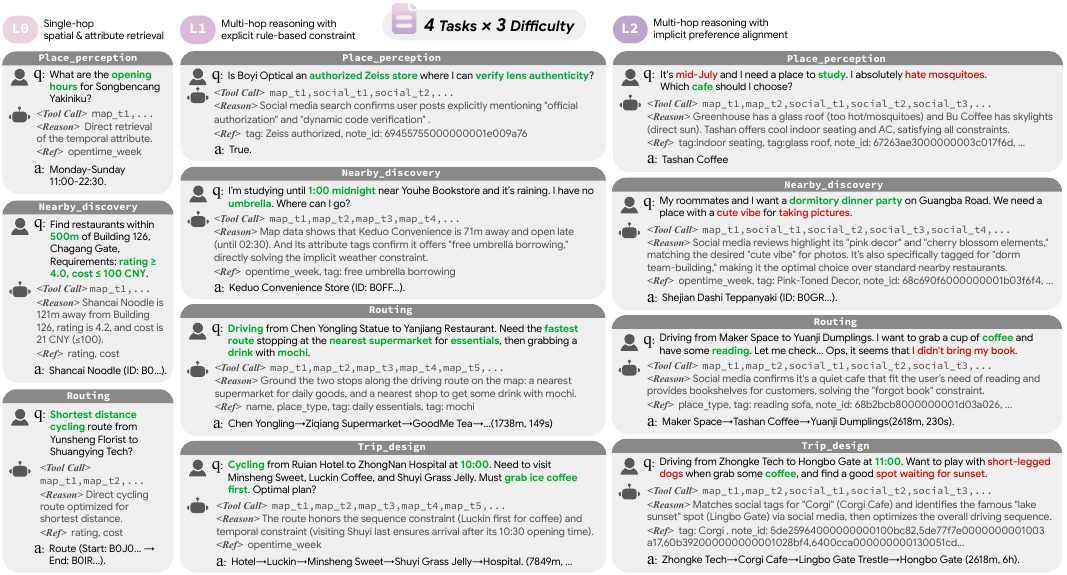}
  \caption{\textbf{LifePlanner Task System.}
  The benchmark spans four task categories and three difficulty levels. L0 requires a single database retrieval step (green), L1 requires multiple retrieval steps and unconstrained calculation, and L2 further requires implicit constraints (red) to be integrated with retrieved evidence for a globally valid plan.
  }
  \vspace{-10pt}
  \label{fig:task_system}
\end{figure*}

\subsection{Environment}
\textbf{Database for LLM querying.}
\label{subsec:data_prepare}
We build the LifePlanner database through a four-stage pipeline:
\begin{itemize}
    \item \textit{Geospatial backbone construction.} We query the AMap API \cite{amap} and OpenStreetMap (OSM) \cite{osm} for places, road networks, and map attributes in the target region. After cleaning and deduplication, each place is assigned a unique spatial anchor ID. As shown in Fig.~\ref{fig:data_enhance}, the corresponding record contains map attributes and routing information and is subsequently enriched with social-media evidence.
    \item \textit{Social-media evidence collection.} We augment each spatial anchor with locally grounded, unstructured evidence from social media, specifically RedNote \cite{rednote}. For each place, we retrieve up to 20 relevant notes by name search and collect up to 10 comments for each note. Each note is stored with its title, full content, publish time, engagement metrics (likes, collects, shares, and comments), and reference tags; each comment is stored with its publish time, full content, and like count. Posts and comments may contain ambiguous or context-dependent place mentions. Such cases remain in the evidence corpus, requiring agents to resolve them by cross-checking place names, spatial anchors, surrounding text, and other retrieved evidence. After privacy removal and availability filtering, the retained corpus contains approximately 200,000 notes and comments.
    \item \textit{Attribute distillation.} We convert the raw social-media corpus into auxiliary structured attributes for dataset construction. Gemini-2.5-Flash is used to extract static venue properties as tags, such as offerings and facilities, and dynamic properties, including long-term impressions (e.g., ambiance and service quality) and short-term states (e.g., new arrivals or temporary closures). As shown in Fig.~\ref{fig:data_enhance}, these attributes are merged into the corresponding spatial anchor record. The distilled dynamic memory fields are used only for case construction and are not available to evaluated agents.
    \item \textit{Access separation for evaluation.} During evaluation, agents must use the MCP tools to retrieve and interpret the original posts and comments associated with candidate places; the distilled dynamic memory fields are not included in their observations. The resulting evidence retains the ambiguity, redundancy, and irrelevant information of the raw social-media corpus.
\end{itemize}

\noindent \textbf{Toolbox for LLM interaction.}
We implement an MCP server \cite{mcp} that emulates how users consult map services and social-media platforms in real life. 
Its core functions are retrieval tools, which allow the LLM to search for target places and obtain corresponding information by ID, name, tag, or free-form description \cite{gao2023retrieval}. We also provide basic geo-spatial analysis tools, such as neighborhood identification.
During evaluation, the LLM specifies a tool-use intent and a structured query, which our tool gateway routes to the appropriate backend tool and returns the resulting feedback. 
Finally, the LLM should return both the final answer ${a}$ and its tool-use chain $\tau=\{(c_t,o_t)\}_{t=1}^{T}$, where $c_t$ and $o_t$ denote the $t$-th tool call and its feedback. Evaluation scores both ${a}$ and $\tau$. Further details are provided in the Appendix.

\subsection{Evaluation Protocol}
\noindent \textbf{Task Taxonomy.}
LifePlanner defines tasks along the scope of spatial decision-making, progressively moving from point-level understanding to area-level itinerary planning:
\begin{itemize}
    \item \textit{Place Perception}: identify, verify, or compare properties of specific places.
    \item \textit{Nearby Discovery}: find suitable candidates around a given spatial context.
    \item \textit{Routing}: identify and integrate optimal waypoints along an active route under specific en-route constraints.
    \item \textit{Trip Design}: construct comprehensive, multi-stop itineraries that jointly satisfy overarching user goals, personal preferences, and spatiotemporal constraints.
\end{itemize}

\noindent \textbf{Difficulty Levels.}
Each task category is stratified by the operations required to solve it:
\begin{itemize}
    \item \textit{L0}: single-step lookup over geospatial data.
    \item \textit{L1}: multi-step retrieval and computation, such as candidate comparison, route checking, or attribute aggregation.
    \item \textit{L2}: hybrid reasoning over retrieved evidence, \textit{implicit constraints, costs, and preferences} to select the best final plan.
\end{itemize}
Note that \textit{Trip Design} starts from L1 because even its simplest form requires multi-step planning. As shown in Fig.~\ref{fig:task_system}, this yields 11 scenario-complexity combinations, enabling systematic analysis across both task diversity and planning complexity.

\noindent \textbf{Evaluation Case Generation.}
We generate evaluation cases with an LLM-based data engine. Given a task-structure template and a candidate answer $a^\star$, the engine has full access to the environment, including map records, raw social-media evidence, condensed dynamic attributes (Sec.~\ref{subsec:data_prepare}), and tools. It first derives a valid solution path by selecting a set of tools, identifying supporting information, and verifying answer uniqueness; it then formulates a query $q$ by injecting semantic and spatial constraints that make this path $\tau^\star$ necessary.
We manually cross-validate each case for solvability, reference correctness, and answer uniqueness. 
Note that the tested LLM receives only the final query $q$ and must reconstruct the hidden solution path through step-by-step database and tool exploration.

\begin{table*}
\centering
\resizebox{\textwidth}{!}{
    \begin{tabular}{l ccc ccc ccc cc cc}
    \toprule
    \multirow{2}{*}{\textbf{Model}} & \multicolumn{3}{c}{\textbf{Place Perception}} & \multicolumn{3}{c}{\textbf{Nearby Discovery}} & \multicolumn{3}{c}{\textbf{Routing}} & \multicolumn{2}{c}{\textbf{Trip}} & \multicolumn{2}{c}{\textbf{Overall}} \\
    \cmidrule(lr){2-4} \cmidrule(lr){5-7} \cmidrule(lr){8-10} \cmidrule(lr){11-12} \cmidrule(lr){13-14}
    & \textit{L0} & \textit{L1} & \textit{L2} & \textit{L0} & \textit{L1} & \textit{L2} & \textit{L0} & \textit{L1} & \textit{L2} & \textit{L1} & \textit{L2} & \textit{Avg.} & \textit{{\#Tokens}} \\
    \midrule
    \multicolumn{14}{l}{\textbf{Closed-source Models}} \\
    \midrule
    Claude-Sonnet-4.6 & \textbf{99.0} & \underline{87.5} & \underline{91.4} & 74.0 & \textbf{67.3} & \textbf{54.3} & \textbf{100.0} & \underline{46.0} & \textbf{38.7} & \underline{69.8} & \underline{25.7} & \underline{62.7} & 121.8k \\
    Qwen3.6-Plus & \textbf{99.0} & 84.7 & 88.6 & \textbf{87.5} & \underline{61.5} & \underline{48.6} & 98.0 & 44.0 & \underline{35.5} & \underline{69.8} & \underline{25.7} & \textbf{63.3} & 142.3k \\
    GPT-5.4 & 97.0 & 83.3 & 80.0 & 75.0 & 44.2 & 40.0 & 95.0 & 44.0 & 29.0 & 66.0 & \textbf{28.6} & 57.9 & 98.9k \\
    Gemini-3.1-Pro-Preview & \underline{98.0} & 86.1 & 71.4 & 78.9 & 53.9 & \underline{48.6} & \textbf{100.0} & 20.0 & 16.1 & 41.5 & 14.3 & 59.4 & 56.5k \\
    \midrule
    \multicolumn{14}{l}{\textbf{Open-source Models}} \\
    \midrule
    GLM-5.1 & \textbf{99.0} & \textbf{88.9} & \textbf{94.3} & \underline{80.8} & \underline{61.5} & 40.0 & \underline{99.0} & \textbf{56.0} & 32.3 & \textbf{73.6} & \underline{25.7} & 61.5 & 75.2k \\
    Kimi-K2.6 & 96.0 & \textbf{88.9} & 82.9 & 76.0 & \textbf{67.3} & 45.7 & 95.0 & 36.0 & \underline{35.5} & 64.2 & \underline{25.7} & 59.8 & 123.2k \\
    MiniMax-M2.7 & 96.0 & 70.8 & 71.4 & \underline{80.8} & \textbf{67.3} & 42.9 & \underline{99.0} & 12.0 & 12.9 & 39.6 & 11.4 & 56.7 & 65.8k \\
    Qwen3.5-27B & 92.0 & 81.9 & 82.9 & 79.8 & 59.6 & 42.9 & 95.0 & 30.0 & 22.6 & 62.3 & 11.4 & 59.4 & 191.2k \\
    Qwen3.5-35B-A3B & 97.0 & 70.8 & 74.3 & 78.9 & 57.7 & 40.0 & 92.0 & 8.0 & 12.9 & 45.3 & 0.0 & 56.7 & 126.4k \\
    Qwen3.5-9B & 83.0 & 47.2 & 17.1 & 64.4 & 21.2 & 25.7 & 83.0 & 4.0 & 0.0 & 32.1 & 2.9 & 41.4 & 90.7k \\
    \bottomrule
    \end{tabular}
}
\caption{\textit{PR@0.9 in \%} (Pass Rate with $\alpha=0.9$) of LLMs. We mark the best result in \textbf{bold} and the second-best with \underline{underlining}.}
\label{tab:main_results}
\vspace{-10pt}
\end{table*}

\textbf{Metric Design.}
We evaluate both final outcomes and intermediate tool-use processes. The outcome metrics assess whether the final answer satisfies task requirements, while the process metrics assess whether the tool-use trajectory supports the answer efficiently and faithfully.

\noindent\textit{Outcome metrics} to score $a$:
\begin{itemize}
    \item \textit{Attribute Accuracy (AA)}. AA evaluates whether the attributes $x$ of the places $p$ in $a$ (Fig.~\ref{fig:data_enhance}) satisfy the explicit attribute requirements $x_q$ specified in the query $q$, such as a store's opening hours. The score is 1 if all requirements are satisfied, and 0 otherwise:
    \begin{equation}
        AA = \mathbbm{1}[x \models x_q].
    \end{equation}

    \item \textit{Place-set Intersection over Union (PIoU)}. PIoU measures the overlap between the predicted place set $p$ in $a$ and the ground-truth target set $p^\star$ in $a^\star$:
    \begin{equation}
        PIoU = \frac{|p \cap p^\star|}{|p \cup p^\star|}.
    \end{equation}

    \item \textit{Routing Deviation (RD)}. For routing-related tasks, RD evaluates whether the distance/time $v$ of the predicted route in $a$ matches the ground-truth $v^\star$ of $a^\star$. Incorrect waypoints introduce route-cost deviations:
    \begin{equation}
        RD_v = PIoU \cdot \max\left(0, 1 - \frac{|v - v^\star|}{v^\star}\right).
    \end{equation}

    \item \textit{Order Consistency (OC)}. For routing and planning tasks, OC checks whether the places shared by $a$ and $a^\star$ appear in the correct order. Let $r(\cdot)$ be the ground-truth rank function, and let $s=p \cap p^\star$ denote the ordered intersection of $p$ and $p^\star$, with $n=|s|$. We use $D(s)$ to count inverted pairs:
    \begin{equation}
        \begin{aligned}
        D(s) &= \sum_{i<j} \mathbbm{1}[r(s_i) > r(s_j)], \\
        OC &=
        \begin{cases}
        PIoU \cdot \left(1 - \dfrac{D(s)}{\binom{n}{2}}\right), & n \geq 2, \\
        0, & n < 2.
        \end{cases}
        \end{aligned}
    \end{equation}

    \item \textit{Semantic Constraint Satisfaction (SC)} evaluates constraints that cannot be fully captured by rules. Three human-calibrated LLM judges, GPT-5.4, Claude Haiku 4.5, and GLM-5.1, receive $a$, $a^\star$, the grading rubric, and verification tools. Each judge assigns a score in $[0,1]$, and the final score is their average. The judge selection and human-calibration procedure is detailed in Appendix~\ref{sec:judge_selection}.

    \item \textit{Pass Rate (PR)}. PR is the most comprehensive indicator that directly evaluates whether a task is successful. It reports the percentage of cases whose SC exceeds a threshold $\alpha$, and all other calculable indicators above are 1. Because both predicted and reference routes are obtained with the same standardized shortest-path tool, selecting the correct places and visit order normally produces a route structure and cost nearly identical to the ground truth. This consistency supports standardized end-to-end checking, while the component metrics provide diagnostic scores for partially correct outputs.
    
\end{itemize}

\noindent\textit{Process metrics} to score $\tau$:
\begin{itemize}
      \item \textit{Information Grounding (IG)} measures whether the intermediate information $o\in\tau$ covers the required information $\tau^\star$, where $o$ denotes identities such as a place attribute or social-media note according to the case:
        \begin{equation}
            IG = \frac{1}{|\{o\}|} \sum_{\{o\}} \mathbbm{1}[o \in \tau^\star].
        \end{equation}
    
    \item \textit{Tool Efficiency (TE)} penalizes redundant tool use by comparing the number of calls in the reference tool-use set $|\tau^\star|$ with the actual number of calls $|\tau|$:
    \begin{equation}
        TE = \min\left(1, \frac{|\tau^\star|}{|\tau|}\right).
    \end{equation}

    \item \textit{Reasoning Purity (RP)} assesses whether the information and reasoning in $\tau$ are faithful to the database, without hallucination or distortion. We provide $\tau$ and the corresponding database evidence to the same human-calibrated LLM judge ensemble for scoring.

    \item \textit{Token Usage} records computational overhead, including input tokens from prompts, queries, tool feedback, and context, as well as all output tokens generated by the LLM agent.
\end{itemize}

\begin{table}
\centering
\small
\begin{tabular}{lccc}
\toprule
\textbf{Metric} & \textbf{L0} & \textbf{L1} & \textbf{L2} \\
\midrule
\multicolumn{4}{l}{\textbf{Process}} \\
\midrule
\textit{Information Grounding} & 96.5 & 80.4 & 64.6 \\
\textit{Tool Efficiency} & 76.2 & 53.7 & 44.6 \\
\textit{Reasoning Purity} & 96.7 & 89.4 & 86.4 \\
\midrule
\multicolumn{4}{l}{\textbf{Outcome}} \\
\midrule
\textit{Attribute Accuracy} & 96.1 & 86.9 & 80.8 \\
\textit{Place-set IoU} & 90.7 & 73.8 & 55.0 \\
\textit{Routing Deviation} & 97.2 & 69.7 & 47.6 \\
\textit{Order Consistency} & - & 84.7 & 44.7 \\
\textit{Semantic Constraint} & - & 74.0 & 65.2 \\
\midrule
\textbf{\textit{Pass Rate@0.9}} & 89.4 & 57.7 & 40.2 \\
\textit{Token Usage} & 24.6k & 137.5k & 251.1k \\
\bottomrule
\end{tabular}
\caption{Average performance in \% of all LLMs across different task difficulty levels. ``-'' indicates that Order Consistency and Semantic Constraint are not applicable to the single-hop retrieval nature of L0 tasks.}
\label{tab:difficulty_levels}
\vspace{-10pt}

\end{table}

\section{Experiments}
\subsection{Models}
We evaluate LifePlanner on a broad set of representative LLMs, including frontier closed-source models (Claude-Sonnet-4.6~\cite{claude4.6}, Gemini-3.1-Pro-Preview~\cite{gemini3.1}, GPT-5.4~\cite{gpt5.4}, and Qwen3.6-Plus~\cite{qwen3.6plus}) and competitive open-source models (GLM-5.1~\cite{glm5team2026glm5}, Kimi-K2.6~\cite{kimik2.6}, MiniMax-M2.7~\cite{minimaxm2.7}, Qwen3.5-27B, Qwen3.5-35B-A3B, and Qwen3.5-9B). Our goal is to compare foundation models under a controlled agent scaffold rather than optimize a separate agent architecture for each model. Accordingly, every model receives the same system prompt, MCP tools, output schema, and maximum budget of 30 tool calls. This standardized setting isolates model-level differences in evidence acquisition, tool use, and constraint integration.
See the Appendix for further details.
\subsection{Outcome Analysis}

\begin{figure}
  \centering
  \includegraphics[width=\linewidth]{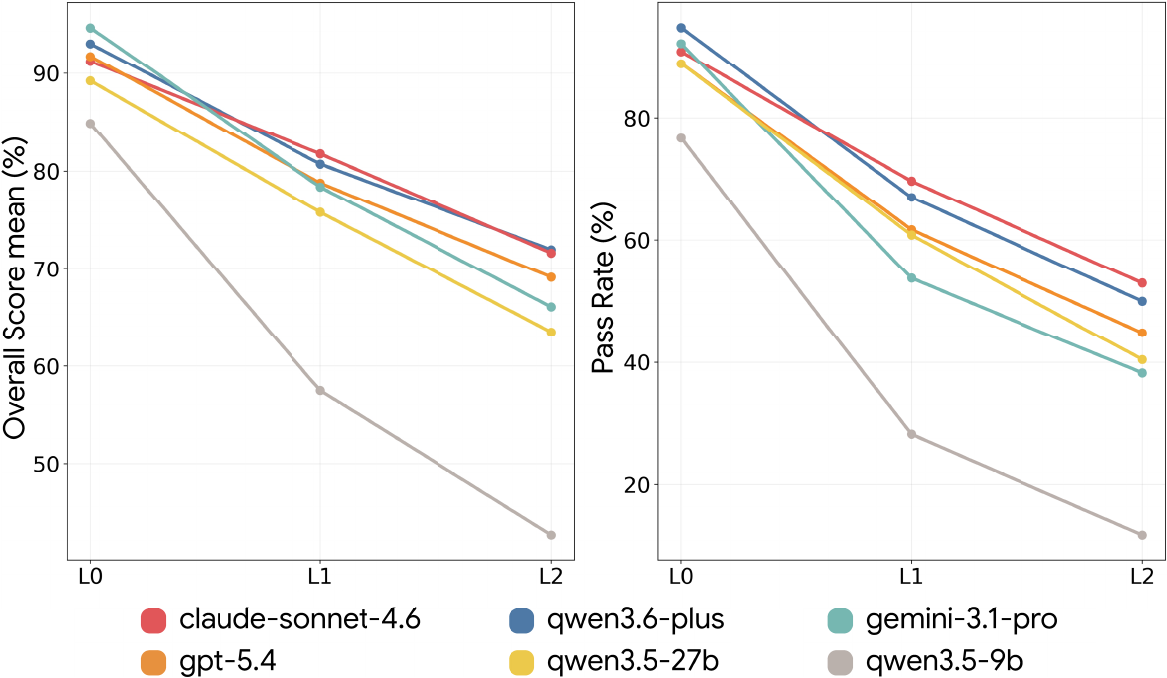}
  \caption{{Average scores} (left) and strict {Pass Rates} (right) of representative models across the three difficulty levels (L0, L1, L2).}
  \label{fig:sc_Avgmetric_PR_compare}
\vspace{-10pt}
\end{figure}

Table~\ref{tab:main_results} presents the comprehensive evaluation results of all models across the 11 tasks. Table~\ref{tab:difficulty_levels} shows model-averaged results across different metrics. We detail our key findings below.

\textbf{Difficulty.} Task difficulty is the most consistent source of performance degradation. From L0 to L2, Table~\ref{tab:main_results}, Table~\ref{tab:difficulty_levels}, and Figure~\ref{fig:sc_Avgmetric_PR_compare} show that the average Pass Rate drops from 89.4\% to 40.2\%, while Token Usage increases from 24.6k to 251.1k. This shows that L2 tasks require longer exploration for multi-hop evidence collection and faithful integration of spatial, semantic, ordering, and implicit constraints. 
However, the outcome metrics of Table~\ref{tab:difficulty_levels} show that current models still perform poorly in evidence retrieval and ordering, and also struggle with constraint understanding and integration, with L2 Semantic Constraint reaching only 65.2\%. Together, these limitations lead to only a 40.2\% success rate on L2 tasks.

\textbf{Tasks.} Performance also varies substantially across task types. In relatively simple place-perception tasks, models need only limited intent understanding and a small number of tool queries; strong models can still achieve over 90\% Pass Rate even at L2. In contrast, trip planning requires models to satisfy multiple retrieval requirements and user constraints simultaneously, with most intermediate steps depending on correct tool use. As a result, even strong models reach only about 25\% Pass Rate on L2 trip design tasks. This demonstrates that our task design effectively distinguishes different dimensions of model capability, while also showing that current models remain far from reliable on complex constrained planning. 

\textbf{Models.} In Table~\ref{tab:main_results}, stronger models generally perform better, but scaling alone does not close the planning gap. Qwen3.6-Plus and Claude-Sonnet-4.6 achieve the highest overall Pass Rates, while GLM-5.1 reaches competitive performance among open-source models. 
Within the same model family, Qwen3.5-27B substantially outperforms Qwen3.5-9B, indicating that scaling is indeed beneficial. However, the ranking is not monotonic with size: GPT-5.4 does not outperform Qwen3.5-27B or GLM-5.1. This suggests that after a basic capability threshold, planning behavior and evidence integration matter more than parameter scale alone.

\textbf{Efficiency.} More exploration is necessary for difficult tasks, but it is not sufficient. Gemini-3.1-Pro-Preview uses the fewest tokens and is highly efficient, yet its early-stopping behavior, i.e., reaching a conclusion before enough evidence has been gathered, hurts performance on difficult routing and trip design cases. Conversely, Qwen3.5-27B consumes 191.2k tokens per task but underperforms GLM-5.1, which uses only 75.2k. Thus, effective planning requires not only collecting more information but also precisely analyzing relevant evidence, making enough well-targeted calls while avoiding redundancy, and integrating constraints into the final answer. This remains a demanding test of an agent's overall capability.

These findings validate our benchmark as a comprehensive platform, highlighting the substantial gap that future LLM agents must bridge: \textit{evidence retrieval in a large database, efficient tool usage, and planning with constraints}.

\begin{figure*}
  \centering
  \includegraphics[width=\linewidth]{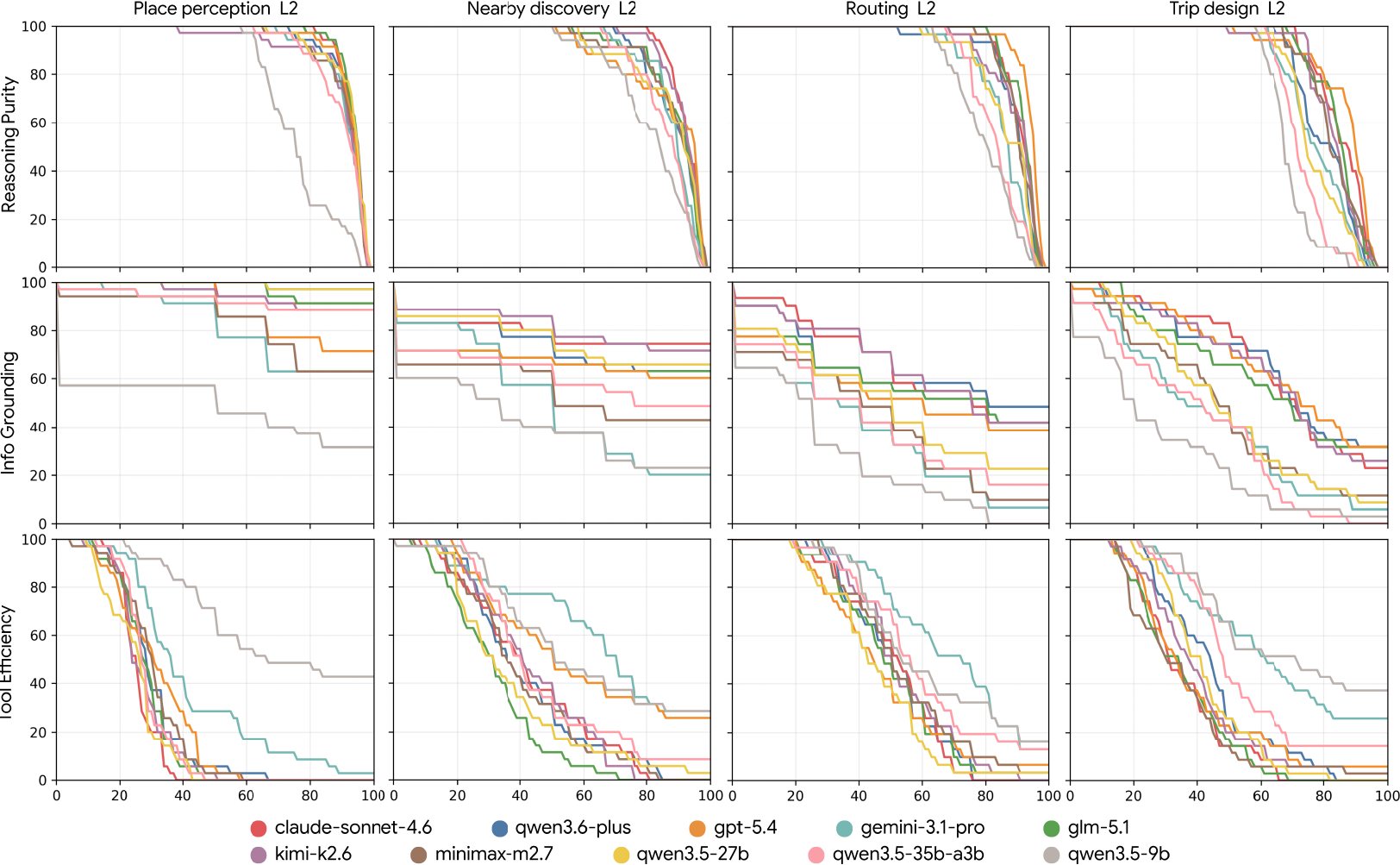}
  \caption{Cross-model exceedance curves for \textbf{Reasoning Purity, Information Grounding, and Tool Efficiency} across different L2 task types. A point $(x,y)$ indicates that $y\%$ of cases achieved a score $\geq x$, where $x$ is expressed as a percentage.}
  \vspace{-10pt}
  \label{fig:L2_RP_IG_TE_compare}
\end{figure*}

\begin{figure}
  \centering
  \includegraphics[width=0.8\linewidth]{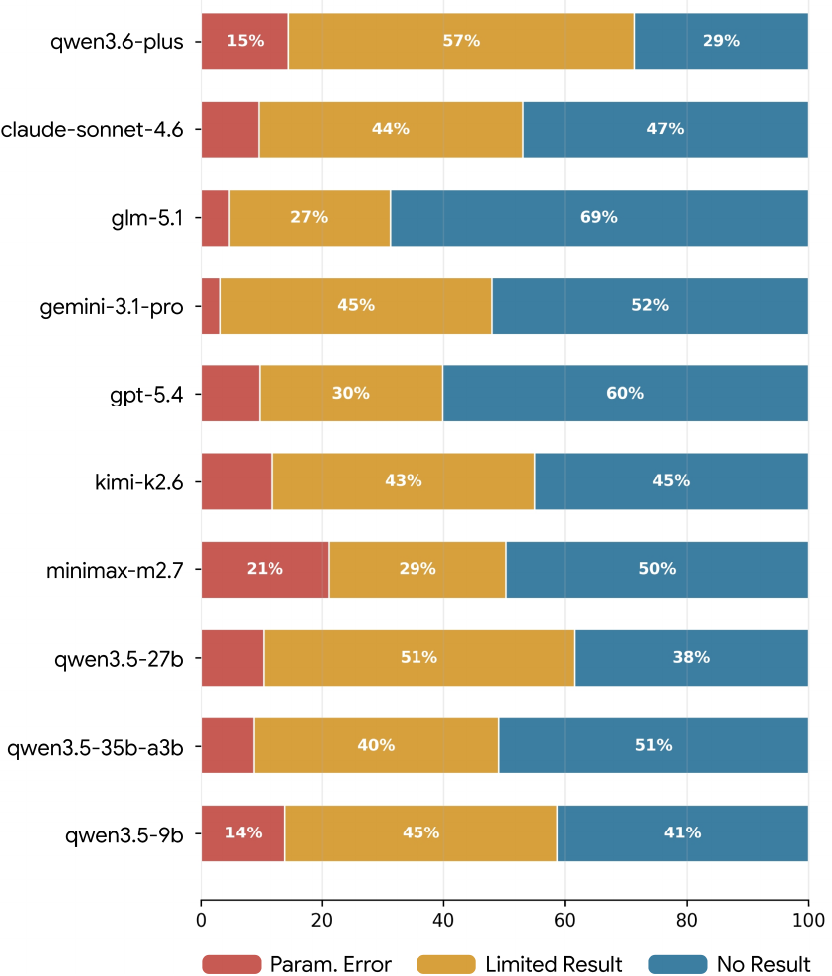}
  \caption{Distribution of tool-call failures caused by parameter errors, truncation due to overly broad conditions, and empty results or errors due to imprecise conditions.}
  \label{fig:tool_fail}
  \vspace{-15pt}
\end{figure}

\subsection{Process Analysis}
\begin{figure}
  \centering
  \includegraphics[width=\linewidth]{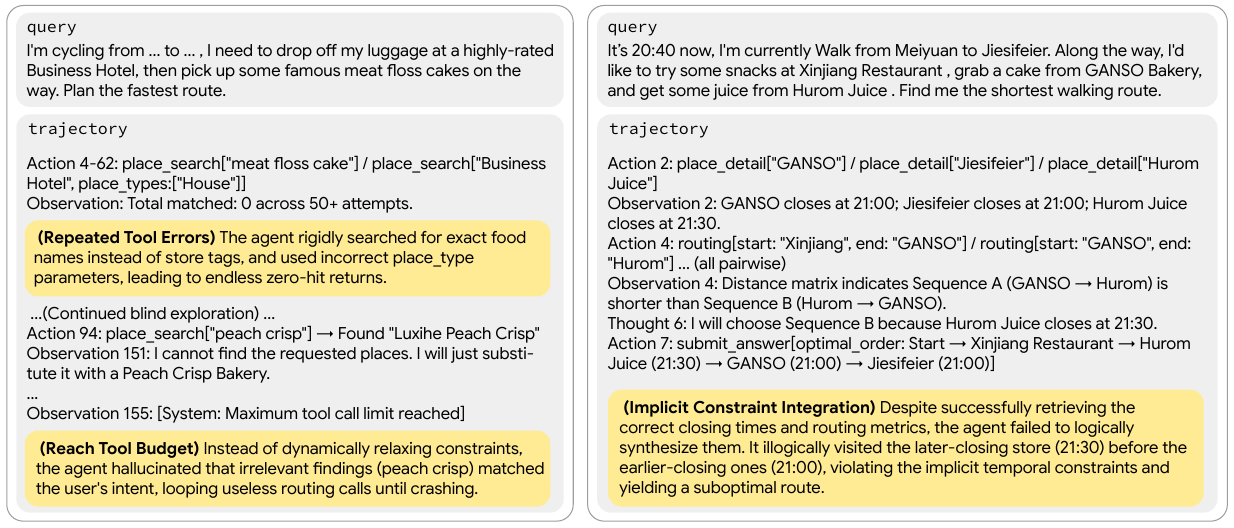}
  \caption{Representative failure cases. Yellow highlights indicate the main errors: (Left) the model repeatedly retrieves the wrong target with an incorrect tag until the tool budget is exhausted; (Right) the model fails to jointly consider constraints such as store closing times when composing the final plan.}
  \label{fig:fail_case}
  \vspace{-20pt}
\end{figure}

We further analyze the process metrics in Table~\ref{tab:difficulty_levels} and Fig.~\ref{fig:L2_RP_IG_TE_compare}, together with the cases in Fig.~\ref{fig:fail_case}, to identify why LLM agents fail during exploration.

\textbf{Faithful reasoning.} Reasoning Purity remains high even on L2 tasks, reaching 86.4\% on average. This indicates that most models can preserve the factual accuracy of retrieved database information without severe hallucination or distortion. However, this ability is not uniform: weaker models, such as Qwen3.5-9B in Fig.~\ref{fig:L2_RP_IG_TE_compare}, show consistently lower Reasoning Purity on L2 tasks, suggesting that faithful use of retrieved evidence still depends on model capability.

\textbf{Incomplete information acquisition.} Information Grounding drops to 64.6\% on L2 tasks, showing that agents frequently make decisions before collecting all required evidence, or exhaust the tool budget without obtaining the necessary information. The gap across models in Fig.~\ref{fig:L2_RP_IG_TE_compare} is substantial. Gemini-3.1-Pro-Preview is trained to terminate early, using the fewest tokens but also retrieving a smaller fraction of required evidence. Conversely, Qwen3.5-27B consumes many more tokens, yet its grounding ratio remains below stronger closed-source models such as Claude-Sonnet-4.6. This confirms that increasing exploration length alone does not solve the problem; agents need more effective evidence analysis and search strategies.

\textbf{Inefficient tool usage.} Tool Efficiency is the weakest process metric, dropping to 44.6\% on L2 tasks. This suggests that models often rely on redundant or poorly targeted tool calls instead of using tools efficiently. 
As shown in Figure~\ref{fig:tool_fail}, tool-use failures are dominated not by an inability to call tools but by imprecise tool use. Only a small fraction of calls fail because of malformed formats or invalid argument types. Most failures instead return limited or empty results, indicating that the model issues queries that are too broad, underspecified, or misaligned with the target evidence. 
Figure~\ref{fig:fail_case} (left) shows a typical failure in which an incorrect tag traps the model in repeated retrieval of irrelevant results until the tool budget is exhausted. In other words, current LLMs generally know how to invoke tools, but still struggle to formulate precise queries and use returned information efficiently.

\textbf{Weak constraint integration.} Even when relevant evidence is retrieved, LLM agents often fail to convert it into a plan satisfying all constraints. Consistent with the Semantic Constraint results of Table~\ref{tab:difficulty_levels}, models may ignore implicit or soft requirements, over-optimize a salient objective such as shortest distance, or misread retrieved evidence when composing the final answer. 
Figure~\ref{fig:fail_case} (right) shows a common case where the model fails to produce a globally reasonable plan because it does not jointly consider the closing-time order of different stores, instead optimizing only for shortest distance. 
Thus, the main bottleneck is not only access to information, but the ability to maintain and integrate multiple constraints in long-horizon reasoning.

Consistent with the outcome analysis, these results suggest that failures are not mainly caused by severe hallucination, but by weak exploration: LLM agents often miss key evidence, issue imprecise tool queries, and fail to turn retrieved information and constraints into reliable plans.

\section{Conclusion}
We present LifePlanner, a realistic geo-spatial planning benchmark that combines structured map data, large-scale social-media evidence, MCP-based tools, and multi-level planning tasks. Our evaluation shows a sharp gap between simple retrieval and complex multi-constraint planning. Failures mainly arise from incomplete evidence acquisition in a large database, inefficient tool usage, and weak constraint integration for joint planning rather than hallucination. Moreover, model scaling and longer reasoning are insufficient without precise tool queries and constrained plan aggregation.

\section*{Limitations}
\textbf{Design limitations}.
Although LifePlanner is designed to scale to massive and dynamic real-world information, the current benchmark has several limitations. First, the social media corpus comes from a single large-scale collection, resulting in a temporally concentrated snapshot. This limits our ability to evaluate how agents handle information conflicts caused by temporal changes.
Second, because the benchmark queries are synthetic, they may state goals and constraints more completely and explicitly than naturally occurring user requests.
Third, the current benchmark focuses on single-turn, multi-hop planning with all constraints given upfront. Extending LifePlanner to multi-turn interactive planning, where user preferences evolve during the conversation, is an important direction for future work.

\noindent \textbf{Potential Risks}.
At present, all external artifacts, including map services, social media data, and model APIs, are used in accordance with their respective access terms and are intended only for research and evaluation; derived resources should not be used for user profiling, commercial redistribution, or purposes beyond the original access conditions.
Because LifePlanner incorporates social media content, privacy protection is a central concern. Although the benchmark is designed for research on planning agents rather than user profiling, raw posts may contain sensitive or personally identifiable information. We therefore remove personal identifiers during preprocessing and retain only task-relevant evidence needed for evaluation. Future dataset releases or extensions should follow platform policies, minimize exposed user content, and apply strict anonymization to prevent re-identification or misuse of personal information.

\bibliography{custom}

\clearpage
\appendix
\label{sec:appendix}

\section{Benchmark Details}
\subsection{Dataset Composition}
In this section, we provide a detailed breakdown of our dataset composition and the underlying data structure. 
Table~\ref{tab:task_distribution} outlines the quantitative distribution of tasks across different categories and difficulty levels. Table~\ref{tab:field_definitions} defines the fields utilized across all task instances with the corresponding descriptions, while Table~\ref{tab:field_coverage} systematically maps which specific fields are triggered under each task family and difficulty tier. Figure~\ref{box:app_task_case} presents a concrete task instance, demonstrating how unstructured user queries, implicit temporal and social-media constraints, and multi-source evidence are structured within our dataset.

\begin{table}
\small
\centering
\caption{Distribution of tasks across different categories and difficulty levels. Note that L0 is not applicable to \textit{Trip} tasks, as they inherently require multi-location retrieval and navigation.}
\label{tab:task_distribution}
\begin{tabular}{lcccc}
\toprule
\textbf{Task Category} & \textbf{L0} & \textbf{L1} & \textbf{L2} & \textbf{Total} \\ 
\midrule
Place Perception   & 100 & 72 & 35 & 207 \\
Nearby Discovery  & 104 & 52 & 35 & 191 \\
Routing & 100 & 50 & 31 & 181 \\
Trip Planning    & --  & 53 & 35 & 88  \\ 
\midrule
\textbf{Total} & \textbf{304} & \textbf{227} & \textbf{136} & \textbf{667} \\ 
\bottomrule
\end{tabular}
\label{tab:app_data_distrib}
\end{table}

\begin{table*}[ht]
\small
\centering
\resizebox{\linewidth}{!}{
\begin{tabularx}{\textwidth}{llX}
\toprule
\textbf{Parent Field} & \textbf{Child Field} & \textbf{Description} \\
\midrule
\texttt{type} & -- & Flag indicating task subtype and difficulty level. (e.g., routing\_L1, trip\_L2) \\
\midrule
\texttt{question} & -- & Natural-language user query presented to the benchmark agent. \\
\midrule
\texttt{answer} & (\textit{Self}) & Root payload object or array containing the final task results or scalar answer. \\
 & \texttt{name} & Name of a returned place, start/end point, waypoint, or itinerary stop. \\
 & \texttt{place\_id} & Map identifier corresponding to the returned location. \\
 & \texttt{reason} & Natural-language explanation for a verification, comparison, route selection, or stop choice. \\
 & \texttt{start} / \texttt{end} & Origin and destination place for route or trip. \\
 & \texttt{vehicle} & Travel mode used for routing (e.g., walking, cycling, driving). \\
 & \texttt{mode} & Route optimization target (e.g., time, distance). \\
 & \texttt{total\_distance\_m} & Total route distance in meters. \\
 & \texttt{total\_time\_s} & Total route time in seconds. \\
 & \texttt{waypoints} & List of intermediate stops used for routing or candidate stops used for trip planning. \\
 & \texttt{optimal\_order} & Visit order of places in the waypoints list. \\
\midrule
\texttt{ref} & (\textit{Self}) & Evidence container attached to an answer or stop to support the decision. \\
 & \texttt{attribute\_key} & List of supporting database fields or map tag values used as evidence. \\
 & \texttt{note\_id} & List of note IDs cited when decisions rely on social-media evidence. \\
 & \texttt{implicit\_logic} & Flag indicating whether the selection depended on implicit or subjective reasoning. \\
\bottomrule
\end{tabularx}
}
\caption{Field definitions for query instances across different tasks. Nested fields are logically grouped under their parent objects.}
\label{tab:field_definitions}
\end{table*}

\begin{table*}[ht]
\small
\centering
\begin{tabularx}{\textwidth}{llX}
\toprule
\textbf{Task Family} & \textbf{Level} & \textbf{Included Fields} \\
\midrule
\multirow{3}{*}{Nearby} & L0 & \texttt{type}, \texttt{question}, \texttt{answer}, \texttt{name}, \texttt{place\_id}, \texttt{reason}, \texttt{ref}, \texttt{attribute\_key} \\
 & L1 & \texttt{type}, \texttt{question}, \texttt{answer}, \texttt{name}, \texttt{place\_id}, \texttt{reason}, \texttt{ref}, \texttt{attribute\_key}, \texttt{note\_id} \\
 & L2 & \texttt{type}, \texttt{question}, \texttt{answer}, \texttt{name}, \texttt{place\_id}, \texttt{reason}, \texttt{ref}, \texttt{attribute\_key}, \texttt{note\_id}, \texttt{implicit\_logic} \\
\midrule
\multirow{3}{*}{Place} & L0 & \texttt{type}, \texttt{question}, \texttt{answer}, \texttt{ref}, \texttt{attribute\_key}, \texttt{note\_id} \\
 & L1 & \texttt{type}, \texttt{question}, \texttt{answer}, \texttt{reason}, \texttt{ref}, \texttt{attribute\_key}, \texttt{note\_id} \\
 & L2 & \texttt{type}, \texttt{question}, \texttt{answer}, \texttt{reason}, \texttt{ref}, \texttt{attribute\_key}, \texttt{note\_id}, \texttt{implicit\_logic}\\
\midrule
\multirow{3}{*}{Routing} & L0 & \texttt{type}, \texttt{question}, \texttt{answer}, \texttt{start} / \texttt{end}, \texttt{name}, \texttt{place\_id}, \texttt{vehicle}, \texttt{mode}, \texttt{total\_distance\_m}, \texttt{total\_time\_s}, \texttt{reason} \\
 & L1 & \texttt{type}, \texttt{question}, \texttt{answer}, \texttt{start} / \texttt{end}, \texttt{waypoints}, \texttt{optimal\_order}, \texttt{name}, \texttt{place\_id}, \texttt{vehicle}, \texttt{mode}, \texttt{total\_distance\_m}, \texttt{total\_time\_s}, \texttt{reason}, \texttt{ref}, \texttt{attribute\_key} \\
 & L2 & \texttt{type}, \texttt{question}, \texttt{answer}, \texttt{start} / \texttt{end}, \texttt{waypoints}, \texttt{optimal\_order}, \texttt{name}, \texttt{place\_id}, \texttt{vehicle}, \texttt{mode}, \texttt{total\_distance\_m}, \texttt{total\_time\_s}, \texttt{reason}, \texttt{ref}, \texttt{attribute\_key}, \texttt{note\_id}, \texttt{implicit\_logic} \\
\midrule
\multirow{2}{*}{Trip} & L1 & \texttt{type}, \texttt{question}, \texttt{answer}, \texttt{start} / \texttt{end}, \texttt{waypoints}, \texttt{optimal\_order}, \texttt{name}, \texttt{place\_id}, \texttt{vehicle}, \texttt{mode}, \texttt{total\_distance\_m}, \texttt{total\_time\_s}, \texttt{reason}, \texttt{ref}, \texttt{attribute\_key} \\
 & L2 & \texttt{type}, \texttt{question}, \texttt{answer}, \texttt{start} / \texttt{end}, \texttt{waypoints}, \texttt{optimal\_order}, \texttt{name}, \texttt{place\_id}, \texttt{vehicle}, \texttt{mode}, \texttt{total\_distance\_m}, \texttt{total\_time\_s}, \texttt{reason}, \texttt{ref}, \texttt{attribute\_key}, \texttt{note\_id}, \texttt{implicit\_logic} \\
\bottomrule
\end{tabularx}
\caption{Field coverage by task family and difficulty level. Raw instance fields have been mapped to the simplified ontology defined in Table~\ref{tab:field_definitions}. Note that \textit{Trip} lacks an L0 configuration due to its inherent complexity.}
\label{tab:field_coverage}
\end{table*}

\subsection{Tool Schema}
In this section, we detail the comprehensive toolset provided to the agents, implemented via the Model Context Protocol (MCP). To accurately simulate human-like progressive exploration, the \textbf{Standard Agent Toolset} (Table~\ref{tab:standard_tools}) is designed to return lightweight, granular initial results. This forces the evaluated agent to actively invoke drill-down tools (e.g., \texttt{place\_detail}, \texttt{note\_detail}) to acquire further insights, mirroring authentic human search behavior. 

Conversely, to maximize dataset generation efficiency and prevent any omission of complex constraints, the \textbf{Data Engine Agent Toolset} (Table~\ref{tab:data_engine_tools}) features enhanced tools. These tools return comprehensive records upfront—including social memory, temporal attributes, and tags—and provide specialized tools for random sampling and multi-stop permutations. Note that the descriptions and parameter explanations listed in both tables represent the exact, raw schema descriptions injected into the agents' prompts.

\begin{table*}[ht]
\small
\centering
\begin{tabularx}{\textwidth}{>{\ttfamily}l >{\raggedright\arraybackslash}X >{\raggedright\arraybackslash}X}
\toprule
\textnormal{\textbf{Tool Name}} & \textbf{Description} & \textbf{Parameters} \\
\midrule
\multicolumn{3}{c}{\cellcolor{gray!20}\textbf{Map Domain}} \\
\midrule
place\_search & Search places by name with lightweight POI output (\texttt{place\_id}, \texttt{name}, \texttt{type}, \texttt{rating}, \texttt{cost}). Supports type, rating, cost, and geo-tag filtering. & 
\textbf{query\_name}: REQUIRED: Place name in local language
\newline 
\textbf{place\_types}: POI types list \newline 
\textbf{min\_rating} / \textbf{max\_rating}: Min/Max rating (1-5) \newline 
\textbf{min\_cost} / \textbf{max\_cost}: Min/Max cost per capita (CNY) \newline 
\textbf{geo\_tag}: Filter by tag/name keywords list in local language, matches if ANY keyword hits \\
\midrule
place\_detail & Get full POI information for one place, including name, type, address, hours, rating, cost, tags, and alias. & 
\textbf{place\_id}: POI place\_id (from place\_search or nearby\_search) \\
\midrule
nearby\_search & Search nearby places around a POI or along a route buffer. Standard agent version returns a lightweight POI list with distance. & 
\textbf{center}: Center: POI place\_id (str) OR Route object \{'start': 'ID', 'end': 'ID', 'vehicle': 'driving'\} \newline 
\textbf{radius\_meters}: Search radius/buffer in meters (default 100) \newline 
\textbf{place\_types}: POI types list \newline 
\textbf{min\_rating} / \textbf{max\_rating}: Min/Max rating (1-5) \newline 
\textbf{min\_cost} / \textbf{max\_cost}: Min/Max cost per capita (CNY) \newline 
\textbf{geo\_tag}: Filter by tag/name keywords list in local language, matches if ANY keyword hits \\
\midrule
routing & Calculate an optimal route between two POIs and return total distance/time, geometry, and turn instructions. & 
\textbf{start\_place\_id}: Start POI place\_id \newline 
\textbf{end\_place\_id}: End POI place\_id \newline 
\textbf{avoid\_way\_ids}: Way IDs to avoid (ways.gid) \newline 
\textbf{mode}: Route mode: 'time' (shortest time) or 'distance' (shortest distance) \newline 
\textbf{vehicle}: Travel mode: 'walking', 'cycling', or 'driving' \\
\midrule
submit\_answer & Submit the final structured JSON answer. The evaluated agent must call this last instead of printing plain JSON. & 
\textbf{result}: Your final answer as a JSON object or array following the required format \\
\midrule
\multicolumn{3}{c}{\cellcolor{gray!20}\textbf{Social Media Domain}} \\
\midrule
search\_notes & Search social media posts by keyword logic and return a lightweight post list (\texttt{note\_id}, title, time, counts). & 
\textbf{keyword}: Search keyword; multiple keywords must be separated by 'AND'/'OR'/'NOT' logic \\
\midrule
note\_detail & Get full post details for one social media note, including description and comments. & 
\textbf{note\_id}: Post ID (from search\_notes result) \\
\bottomrule
\end{tabularx}
\caption{The Standard Agent Toolset. Tools are designed to be lightweight, requiring the evaluated agent to execute multi-step reasoning and drill-down tool invocations to fulfill complex queries.}
\label{tab:standard_tools}
\end{table*}

\begin{table*}[ht]
\small
\centering
\begin{tabularx}{\textwidth}{>{\ttfamily}l >{\raggedright\arraybackslash}X >{\raggedright\arraybackslash}X}
\toprule
\textnormal{\textbf{Tool Name}} & \textbf{Description} & \textbf{Parameters} \\
\midrule
\multicolumn{3}{c}{\cellcolor{gray!20}\textbf{Map Domain}} \\
\midrule
place\_search & Search places by name with enhanced filters and return full POI records. Data Engine version can filter by both static geo tags and social-memory keywords. & 
\textbf{query\_name}: REQUIRED: Place name in local language \newline 
\textbf{place\_types}: POI types list, separated by commas (,) \newline 
\textbf{min\_rating} / \textbf{max\_rating}: Min/Max rating (1-5) \newline 
\textbf{min\_cost} / \textbf{max\_cost}: Min/Max cost per capita \newline 
\textbf{geo\_tag}: Filter by tag/name keywords list \newline 
\textbf{social\_mem}: Filter by social insights in local language\\
\midrule
nearby\_search & Search nearby places around a POI or along a route buffer and return full POI records, including \texttt{address}, \texttt{opentime\_week}, \texttt{tag}, \texttt{alias}, \texttt{long\_mem}, and \texttt{short\_mem}. & 
\textbf{center}: Center: POI place\_id (str) OR Route object \{'start': 'ID', 'end': 'ID', 'vehicle': 'driving'\} \newline 
\textbf{radius\_meters}: Search radius/buffer in meters \newline 
\textbf{place\_types}: POI types list, separated by commas \newline 
\textbf{min\_rating} / \textbf{max\_rating}: Min/Max rating (1-5) \newline 
\textbf{min\_cost} / \textbf{max\_cost}: Min/Max cost per capita \newline 
\textbf{geo\_tag}: Filter by tag/name keywords list \newline 
\textbf{social\_mem}: Filter by long\_mem/short\_mem keywords (social media insights) in local language \\
\midrule
routing & Calculate an optimal route between two POIs and return total distance/time, geometry, and turn instructions. & 
\textbf{start\_place\_id}: Start POI place\_id \newline 
\textbf{end\_place\_id}: End POI place\_id \newline 
\textbf{avoid\_way\_ids}: Way IDs to avoid (ways.gid) \newline 
\textbf{mode}: Route mode: 'time' or 'distance' \newline 
\textbf{vehicle}: Travel mode: 'walking', 'cycling', or 'driving' \\
\midrule
sample\_place & Randomly sample POIs from the map database for prompt initialization. Returns full POI information including memory fields. & 
\textbf{num\_samples}: Number of random POIs to sample (typically 1-50) \\
\midrule
multi\_stop\_routing & Enumerate and score route permutations for multi-stop trip planning. Supports open trips and fixed start/end settings. & 
\textbf{stop\_ids}: List of POI place\_ids to visit (typically 2-6 stops) \newline 
\textbf{start\_id}: Fixed start place\_id (must be in stop\_ids) \newline 
\textbf{end\_id}: Fixed end place\_id (must be in stop\_ids) \newline 
\textbf{vehicle}: Travel mode: 'walking', 'cycling', 'driving' \newline 
\textbf{mode}: Optimize: 'time' or 'distance' \\
\midrule
submit\_answer & Submit the final structured JSON answer generated by the Data Engine agent. & 
\textbf{result}: Your final answer as a JSON object or array following the required format \\
\midrule
\multicolumn{3}{c}{\cellcolor{gray!20}\textbf{Social Media Domain}} \\
\midrule
search\_notes & Search social media posts and return full post information with comments. Required when generated data cites social-memory evidence. & 
\textbf{keyword}: Search keyword; multiple keywords must be separated by 'AND'/'OR'/'NOT' logic \\
\bottomrule
\end{tabularx}
\caption{The Data Engine Agent Toolset. Enhanced MCP tools directly return highly dense and comprehensive records to ensure complete information retrieval during dataset generation without information loss.}
\label{tab:data_engine_tools}
\end{table*}

\subsection{Database Schema}
In this section, we delineate the schema of the underlying environment databases. It is important to note that the fields detailed in Table~\ref{tab:map_db_schema} (Map Database) and Table~\ref{tab:social_db_schema} (Social Media Database) only represent the attributes explicitly retrievable by the agents via the MCP tools. Internal environmental data essential for system operation—such as the topological geometries of road networks (ways/nodes) used for routing, or the tokenized texts and vector embeddings utilized for social media full-text search (\texttt{notes\_fts})—are abstracted away and intentionally not exposed to the agents.

\begin{table*}[ht]
\small
\centering
\begin{tabularx}{\textwidth}{>{\ttfamily}l >{\raggedright\arraybackslash}X >{\raggedright\arraybackslash}X}
\toprule
\textnormal{\textbf{Returned Key}} & \textbf{Exposed by Tool(s)} & \textbf{Explanation} \\
\midrule
osm\_id & \texttt{place\_search}, \texttt{nearby\_search}, \texttt{sample\_place}, \texttt{place\_detail} & Primary POI identifier. \\
\midrule
name & \texttt{place\_search}, \texttt{nearby\_search}, \texttt{sample\_place}, \texttt{place\_detail} & POI name. \\
\midrule
place\_type & \texttt{place\_search}, \texttt{nearby\_search}, \texttt{sample\_place}, \texttt{place\_detail} & Categorization of the POI (e.g., Auto Service|Dedicated Charging Station). \\
\midrule
address & \texttt{place\_search} (DE), \texttt{nearby\_search} (DE), \texttt{sample\_place}, \texttt{place\_detail} & Physical address of the POI. \\
\midrule
opentime\_week & \texttt{place\_search} (DE), \texttt{nearby\_search} (DE), \texttt{sample\_place}, \texttt{place\_detail} & Weekly business opening hours, if available. \\
\midrule
rating & \texttt{place\_search}, \texttt{nearby\_search}, \texttt{sample\_place}, \texttt{place\_detail} & User rating score. \\
\midrule
cost & \texttt{place\_search}, \texttt{nearby\_search}, \texttt{sample\_place}, \texttt{place\_detail} & Average consumption cost per person. \\
\midrule
tag & \texttt{place\_search} (DE), \texttt{nearby\_search} (DE), \texttt{sample\_place}, \texttt{place\_detail} & Static attribute tags associated with the POI. \\
\midrule
alias & \texttt{place\_search} (DE), \texttt{nearby\_search} (DE), \texttt{sample\_place}, \texttt{place\_detail} & Known aliases or alternate names. \\
\midrule
long\_mem & \texttt{place\_search} (DE), \texttt{nearby\_search} (DE), \texttt{sample\_place} & Long social-memory summary synthesized and attached to the POI. \\
\midrule
short\_mem & \texttt{place\_search} (DE), \texttt{nearby\_search} (DE), \texttt{sample\_place} & Short social-memory summary synthesized and attached to the POI. \\
\midrule
steps[].way\_ids & \texttt{routing} & Underlying routing edge identifiers exposed in step-level route output. \\
\midrule
road\_name & \texttt{routing} & Road names used in step-level route navigation instructions. \\
\bottomrule
\end{tabularx}
\caption{Retrievable fields from the Map Database. Fields marked with (DE) are exclusively exposed upfront to the Data Engine Agent during the initial search.}
\label{tab:map_db_schema}
\end{table*}

\begin{table*}[ht]
\small
\centering
\begin{tabularx}{\textwidth}{>{\ttfamily}l >{\raggedright\arraybackslash}X >{\raggedright\arraybackslash}X}
\toprule
\textnormal{\textbf{Returned Key}} & \textbf{Exposed by Tool(s)} & \textbf{Explanation} \\
\midrule
note\_id & \texttt{search\_notes}, \texttt{note\_detail} & Primary identifier for the social media post. \\
\midrule
title & \texttt{search\_notes}, \texttt{note\_detail} & Title of the post. \\
\midrule
desc & \texttt{search\_notes} (DE), \texttt{note\_detail} & Full description and textual content of the post. \\
\midrule
time & \texttt{search\_notes}, \texttt{note\_detail} & Publication timestamp of the post. \\
\midrule
liked\_count & \texttt{search\_notes}, \texttt{note\_detail} & Number of likes received by the post. \\
\midrule
collected\_count & \texttt{search\_notes}, \texttt{note\_detail} & Number of times the post has been saved or favorited. \\
\midrule
comment\_count & \texttt{search\_notes}, \texttt{note\_detail} & Total number of comments under the post. \\
\midrule
share\_count & \texttt{search\_notes} (DE), \texttt{note\_detail} & Number of times the post has been shared. \\
\midrule
tag\_list & \texttt{search\_notes} (DE), \texttt{note\_detail} & Comma-separated list of tags or hashtags attached to the post. \\
\midrule
comment\_id & \texttt{search\_notes} (DE), \texttt{note\_detail} & Unique identifier for an individual comment. \\
\midrule
comments[].time & \texttt{search\_notes} (DE), \texttt{note\_detail} & Timestamp of when the specific comment was posted. \\
\midrule
comments[].content & \texttt{search\_notes} (DE), \texttt{note\_detail} & The text content of the specific comment. \\
\midrule
comments[].like\_count & \texttt{search\_notes} (DE), \texttt{note\_detail} & Number of likes received by the specific comment. \\
\bottomrule
\end{tabularx}
\caption{Retrievable fields from the Social Media Database. The Data Engine Agent (DE) retrieves detailed comment data directly from the search tool to ensure data integrity during generation.}
\label{tab:social_db_schema}
\end{table*}

\section{Evaluation Details}
\subsection{Metric Applicability}
Table~\ref{tab:metric_matrix} details the applicability of each evaluation metric across different task categories and difficulty levels within the benchmark.

\subsection{Judge Selection and Human Calibration}
\label{sec:judge_selection}
We select the LLM judges through a human-calibrated screening procedure. We first ask the flagship model from each of the Claude, GPT, and Gemini families to independently score the subjective dimensions of responses produced by an evaluated model from a separate model family. We then select the 100 cases with the largest variance across the three preliminary scores to form a challenging calibration subset. Human annotators assign reference scores to these cases using the same grading rubric and verification evidence.

We subsequently evaluate a broader pool of candidate judge models on this subset and compare their scores with the human references. GPT-5.4, Claude Haiku 4.5, and GLM-5.1 show the closest overall agreement with the human scores and are therefore selected as the final judge ensemble. The reported Semantic Constraint Satisfaction and Reasoning Purity scores are averaged over these three judges.

\subsection{Hyperparameters}
To minimize variance and ensure strict evaluation reproducibility, the temperature parameter for all evaluator and judge models is consistently set to 0. The temperature for the data engine agent (during task generation) is set to 0.7 to encourage generative diversity. To isolate model differences under a controlled scaffold, all evaluated models use the same system prompt, MCP tool schemas, output requirements, and maximum budget of 30 tool invocations per task instance.

\begin{table*}[t]
\centering
\begin{tabular}{l ccc ccc ccc cc}
\toprule
\multirow{2}{*}{\textbf{Metric}} & \multicolumn{3}{c}{\textbf{\makecell{Place\\Perception}}} & \multicolumn{3}{c}{\textbf{\makecell{Nearby\\Discovery}}} & \multicolumn{3}{c}{\textbf{Routing}} & \multicolumn{2}{c}{\textbf{\makecell{Trip\\Design}}} \\
\cmidrule(lr){2-4} \cmidrule(lr){5-7} \cmidrule(lr){8-10} \cmidrule(lr){11-12}
& \textit{L0} & \textit{L1} & \textit{L2} & \textit{L0} & \textit{L1} & \textit{L2} & \textit{L0} & \textit{L1} & \textit{L2} & \textit{L1} & \textit{L2} \\
\midrule
Attr Accuracy       & \checkmark & \checkmark & \checkmark &            &            &            & \checkmark & \checkmark & \checkmark & \checkmark & \checkmark \\
Place-set IoU       &            &            &            & \checkmark & \checkmark & \checkmark & \checkmark & \checkmark & \checkmark & \checkmark & \checkmark \\
Routing Deviation   &            &            &            &            &            &            & \checkmark & \checkmark & \checkmark & \checkmark & \checkmark \\
Order Consistency   &            &            &            &            &            &            &            &            &            & \checkmark & \checkmark \\
Semantic Constraint &            & \checkmark & \checkmark &            & \checkmark & \checkmark &            & \checkmark & \checkmark & \checkmark & \checkmark \\
\midrule
Info Grounding      & \checkmark & \checkmark & \checkmark & \checkmark & \checkmark & \checkmark & \checkmark & \checkmark & \checkmark & \checkmark & \checkmark \\
Tool Efficiency     & \checkmark & \checkmark & \checkmark & \checkmark & \checkmark & \checkmark & \checkmark & \checkmark & \checkmark & \checkmark & \checkmark \\
Reasoning Purity    & \checkmark & \checkmark & \checkmark & \checkmark & \checkmark & \checkmark & \checkmark & \checkmark & \checkmark & \checkmark & \checkmark \\
Token Usage    & \checkmark & \checkmark & \checkmark & \checkmark & \checkmark & \checkmark & \checkmark & \checkmark & \checkmark & \checkmark & \checkmark \\
\bottomrule
\end{tabular}
\caption{Applicability matrix of evaluation metrics across different task types and difficulty levels.}
\label{tab:metric_matrix}
\end{table*}

\section{Prompt List}
In this section, we provide the complete system prompts used to construct the autonomous agents in our framework. Figure~\ref{box:agent_prompt} presents the foundational system prompt for the evaluated benchmark agent, which outlines its persona, task objectives, and reasoning constraints when navigating the environment. Figure~\ref{box:data_agent_prompt} details the system prompt for the Data Engine Agent, emphasizing its specialized directives for comprehensive data retrieval and implicit constraint synthesis.

\section{AI Usage in Research}

\textbf{Annotation.} We used LLM-based data engines to synthesize evaluation cases, generate candidate answers, and construct reference tool-use paths from task specifications and retrieved evidence. All generated cases were manually checked by the authors for solvability, reference correctness, and answer uniqueness.

\textbf{Evaluation.} We used LLM-as-a-judge models for metrics that require semantic assessment, including Semantic Constraint Satisfaction and Reasoning Purity. The judge ensemble was selected through the human-calibrated screening procedure described in Appendix~\ref{sec:judge_selection}. These judgments were combined with rule-based metrics and verification tools, and the authors reviewed the evaluation design and outputs.

\textbf{Writing.} During the preparation of this work, the authors used ChatGPT to improve language and readability. The authors subsequently reviewed and edited the resulting text as needed. They take full responsibility for the content of the publication.

\clearpage
\onecolumn

\begin{tcblisting}{
    breakable,
    listing only,
    colback=gray!5,
    colframe=black!60, 
    coltitle=white,
    fonttitle=\bfseries\large,
    title=Example for Task Instance,
    boxrule=1pt,
    arc=3pt,
    left=5pt,
    right=5pt,
    top=5pt,
    bottom=5pt,
    width=\textwidth,
    listing options={
        breaklines=true,
        basicstyle=\ttfamily\small
    }
}
{
  "type": "trip_L2",
  "question": "I'm leaving work at Wuhan Zhongke Kaiwu Technology Co., Ltd. and driving to meet a friend at Hongbo Gate. It's 11:02 a.m. now. On the way, I want to stop somewhere to play with some short-legged dogs, and also visit the popular photo spot where I can walk above the lake and waiting for the sunset. Plan the quickest driving route."
  ,
  "answer": {
    "start": {
      "name": "Wuhan Zhongke Kaiwu Technology Co., Ltd.",
      "place_id": "B0FFG2DQ5U",
      "selectedreason": "User-specified origin."
    },
    "end": {
      "name": "Hongbo Gate",
      "place_id": "B0LR9NNY1H",
      "reason": "User-specified destination."
    },
    "waypoints": [
      {
        "name": "Corgi & Shiba Inu Pet Cafe (Jiedaokou Branch)",
        "place_id": "B0FFMBUXIP",
        "reason": "Dedicated Corgi and Shiba Inu cafe that verified through social media, meeting the requirements for playing with short-legged dogs. Business hours are 11:00-21:00, arrival time is approximately 11:06, within opening hours.",
        "ref": {
          "note_id": "5de25964000000000100bc82,68fee6b5000000000301305a",
          "attribute_key": "tag:corgi,opentime_week",
          "implicit_logic": "True"
        }
      },
      {
        "name": "Lingbo Gate Trestle",
        "place_id": "B0KANCP2BM",
        "reason": "A popular viewing spot, you can walk on the trestle on the lake and is a popular spot to enjoy the sunset. Social media notes indicate that this is a well-known spot for sunrise/sunset viewing.",
        "ref": {
          "note_id": "6769ff67000000000900fe52,6759819b000000000203527b",
          "implicit_logic": "True"
        }
      }
    ],
    "optimal_order": [
      {
        "name": "Wuhan Zhongke Kaiwu Technology Co., Ltd.",
        "place_id": "B0FFG2DQ5U"
      },
      {
        "name": "Corgi & Shiba Inu Pet Cafe (Jiedaokou Branch)",
        "place_id": "B0FFMBUXIP"
      },
      {
        "name": "Lingbo Gate Trestle",
        "place_id": "B0KANCP2BM"
      },
      {
        "name": "Hongbo Gate",
        "place_id": "B0LR9NNY1H"
      }
    ],
    "vehicle": "driving",
    "mode": "time",
    "total_distance_m": 8618,
    "total_time_s": 726
  }
}
\end{tcblisting}
\captionof{figure}{An example of task instance in LifePlanner.}
\label{box:app_task_case}

\vspace{2em}
\begin{tcblisting}{
    breakable,
    listing only,
    colback=gray!5, 
    colframe=black!60, 
    coltitle=white,
    fonttitle=\bfseries\large,
    title=System Prompt for the Evaluated Agent,
    boxrule=1pt,
    arc=3pt,
    left=5pt,
    right=5pt,
    top=5pt,
    bottom=5pt,
    width=\textwidth,
    listing options={
        breaklines=true,
        basicstyle=\ttfamily\small
    }
}
You are a geographic information assistant. Use the available tools to answer the user's question.

CRITICAL RULES:
1. NEVER hallucinate or fabricate data. ALL reasoning MUST be grounded in tool return results.
2. You MUST call tools to get real data before answering.
3. If your answer references social media insights, you MUST include ALL relevant note_ids in the ref field. 4. Use the EXACT values (names, IDs, distances, times) returned by tools.
5. After completing your analysis, you MUST call submit_answer() to deliver your final answer as a structured JSON object. Do NOT output JSON as plain text - always use submit_answer().


---

You MUST call submit_answer(result=<your_answer>) where <your_answer> is a single JSON object containing ALL of the following fields:
{
    "start": {
      "name": "{NAME}",
      "place_id": "{PLACE_ID}",
      "reason": "Why this place was selected based on criteria."
    },
    "end": {
      "name": "{NAME}",
      "place_id": "{PLACE_ID}",
      "reason": "Why this place was selected based on criteria."
    },
    "waypoints": [
      {
        "name": "{NAME}", 
        "place_id": "{PLACE_ID}", 
        "reason": "Why this place was selected based on criteria.",
        "ref": {
          "note_id": "{NOTE_ID1},{NOTE_ID2},...",
          "attribute_key": "comma-separated keys. tag:keyword,rating,cost",
          "implicit_logic": "True"
        }
      }
    ],
    "optimal_order": [
      {"name": "{NAME}", "place_id": "{PLACE_ID}", "visit_order": 1},
      ...
    ],
    "vehicle": "{VEHICLE}",
    "mode": "{MODE}",
    "total_distance_m": <number>,
    "total_time_s": <number>,
    "constraints_satisfied": ["description of each constraint met"]
  }

IMPORTANT: Pass the ENTIRE object above as the `result` parameter. Do NOT split fields into separate parameters. Use the EXACT field names shown above.


---

Now answer the following question:
[benchmark question text inserted here at runtime]
\end{tcblisting}
\captionof{figure}{The complete system prompt provided to the evaluated agent, defining its reasoning guidelines and tool-use constraints.}
\label{box:agent_prompt}

\vspace{2em}
\begin{tcblisting}{
    breakable,
    listing only,
    colback=gray!5,
    colframe=black!60, 
    coltitle=white,
    fonttitle=\bfseries\large,
    title=System Prompt for the Data Engine Agent,
    boxrule=1pt,
    arc=3pt,
    left=5pt,
    right=5pt,
    top=5pt,
    bottom=5pt,
    width=\textwidth,
    listing options={
        breaklines=true,
        basicstyle=\ttfamily\small
    }
}
You are a Synthetic Data Generation Agent for a Map Assistant Benchmark.
Your goal is to reverse-engineer realistic multi-stop trip planning queries based on the provided map environment data.
Follow the "Core Principles": Groundedness, Diversity, strict JSON format, Uniqueness, output Language, and Mandatory Tool Calls.

**Core Principles:**
1. **Groundedness**: Your generated questions MUST be answerable using routing results from the tools. Do not hallucinate places or routes not present in the tool output.
2. **Diversity**: 
   - Vary the phrasing (e.g., "Plan a trip visiting...", "I need to go to these places...", "What's the best order to visit...").
   - Vary vehicle types: walking, cycling, driving. Vary constraints: time windows, ordering, efficiency.
   - **CRITICAL**: For L2, diversify the tag/mem conditions. Explore various categories and attributes.
3. **Format**: Output strictly in the specified JSON format. No markdown fencing (```json) outside the JSON block.
4. **Necessity of Uniqueness**: 
   - The selected places must be the ONLY ones matching the criteria in the entire map region.
   - Verify uniqueness using `place_search` with strict filters.
5. **Language**: ALL generated content MUST be in ENGLISH, except for place names and social media content (which should remain in their original language).
6. **Mandatory Tool Calls**:
   - You MUST call `multi_stop_routing()` to get ACTUAL total_distance_m and total_time_s values. DO NOT fabricate these numbers.
   - The `total_distance_m` and `total_time_s` in your final output MUST exactly match the tool's return values.
   - Skipping tool calls and guessing route metrics is STRICTLY FORBIDDEN.


**Map Region**: {Current City}. Use local language of current city for place names and queries in this region.


---

# CURRENT TASK: L2: Mixed Reasoning + Dynamic Waypoint Selection

**Description**: Select waypoints based on tag/mem conditions, then plan optimal route.

# OUTPUT FORMAT
{
  "type": "trip_reasoning",
  "question": "A complex query requiring selection of waypoints based on implicit criteria, then route planning.",
  "answer": {
    "start": {
        "name": "{NAME}", 
        "osm_id": "{OSM_ID}",
        "reason": "Why this place was selected based on criteria.",
        "ref": {
          "note_id": "{NOTE_ID1},{NOTE_ID2},...",
          "attribute_key": "comma-separated keys. tag:keyword,rating,cost",
          "implicit_logic": "True"
        }
      }
    "end": {
        "name": "{NAME}", 
        "osm_id": "{OSM_ID}",
        "reason": "Why this place was selected based on criteria."
        "ref": {
          "note_id": "{NOTE_ID1},{NOTE_ID2},...",
          "attribute_key": "comma-separated keys. tag:keyword,rating,cost",
          "implicit_logic": "True"
        }
      }
    "waypoints": [
      {
        "name": "{NAME}", 
        "osm_id": "{OSM_ID}", 
        "reason": "Why this place was selected based on criteria.",
        "ref": {
          "note_id": "{NOTE_ID1},{NOTE_ID2},...",
          "attribute_key": "comma-separated keys. tag:keyword,rating,cost",
          "implicit_logic": "True"
        }
      }
    ],
    "optimal_order": [
      {"name": "{NAME}", "osm_id": "{OSM_ID}", "visit_order": 1},
      ...
    ],
    "vehicle": "{VEHICLE}",
    "mode": "{MODE}",
    "total_distance_m": <number>,
    "total_time_s": <number>,
    "constraints_satisfied": ["description of each constraint met"]
  }
}

## REF FIELD RULES (MUST follow):
- **tag content**: Write `"tag:{specific_text}"`. Never just `"tag"`.
- **Other DB fields**: Write field name directly, e.g. `"rating"`, `"cost"`, `"opentime_week"`.
- **Multiple sources**: Comma-separate, e.g. `opentime_week"`, `"rating,cost"`.
- **note_id**: REQUIRED when reasoning uses any social media insight (long_mem/short_mem content). Must be real ID(s) from `search_notes()`. Comma-separate multiple IDs.
- The long_mem/short_mem fields are summaries of social posts - always trace back to the original note_id(s).
- **NEVER** write `"attribute_key": "long_mem"` or `"short_mem"`. Find the source note and cite the actual DB fields that informed the decision.
- If no matching note is found via `search_notes()`, discard that place and select a different one.


---

# TASK INSTRUCTIONS

**Note**: The system has provided 10 sampled places above for REFERENCE only.
You MUST call tools (`place_search`, `multi_stop_routing`, `search_notes`) to validate and get actual data.

Choose ONE scenario:

--- SCENARIO A: Full Discovery Trip (No Fixed Points) ---
*Goal*: Find N places matching criteria, then plan optimal route visiting all.

[A1. Define Selection Criteria]:
  Design a multi-constraint question combining 2+ info sources (tag, rating, cost, opentime, social media insights). The locations sampled are used as sources of inspiration.
  Good questions layer MULTIPLE constraints, e.g.:
  - "pet cafe with Corgis" -> geo_tag = "Corgis" + place_type = "Food & Beverages"
  - "affordable KTV with good reviews" -> rating + cost + social notes

[A2. Search and Validate Uniqueness (MANDATORY)]:
  **YOU MUST CALL THIS TOOL - DO NOT SKIP**
  Call `place_search` with criteria in A1.
  **CRITICAL**: Result count must be 3-5 (>=3 required - with only 2 waypoints there is no ordering to optimize, it degenerates to a simple A -> B routing question).
  - If too many (>5): add more constraints.
  - If too few (<3): relax constraints or change criteria.
  These become the waypoints.

[A3. Compute Optimal Order (MANDATORY)]:
  **YOU MUST CALL THIS TOOL - DO NOT SKIP**
  Call `multi_stop_routing(stop_ids=[selected IDs], vehicle={VEHICLE}, mode={MODE})`.
  This returns ALL permutations sorted by total time/distance.
  Extract the ACTUAL `total_distance_m` and `total_time_s` from the tool response.
  Select the first (optimal) route from the result.

[A4. Question Formulation]:
  Construct a persona-based question that implies the criteria.
  Example: "I'm a dog owner. Plan a cafe-hopping route to all pet-friendly coffee shops in the area."
  Do NOT directly mention the tag-use natural language.

[A5. Reference Verification (MANDATORY for social insights)]:
  **IF using social insights (long_mem/short_mem), YOU MUST CALL `search_notes`**
  For each selected waypoint using social insights:
  - Call `search_notes` with relevant keywords.
  - Extract {NOTE_ID}s.

[A6. Final Output]:
  Include `waypoints` with `reason` and `ref`.
  **CRITICAL**: Use the EXACT values from tool response for total_distance_m and total_time_s.

--- SCENARIO B: Constrained Trip (Fixed Start/End, Constraints FORCE Detour) ---
*Goal*: Given fixed start/end, find N intermediate places matching criteria, then plan route where constraints FORCE a different order than the unconstrained optimal.

[B1. Select Fixed Endpoints]:
  From the pre-sampled places below, select {START} and {END} place.

[B2. Define Selection Criteria for Waypoints]:
  Design multi-constraint criteria combining 2+ info sources (tag, rating, cost, opentime, social notes). Each waypoint should require distinct reasoning.
  **CRITICAL**: >=2 intermediate waypoints required - with only 1, the order is trivially start -> waypoint -> end and there is nothing to optimize.

[B3. Search and Validate (MANDATORY)]:
  **YOU MUST CALL THIS TOOL - DO NOT SKIP**
  Call `place_search` with criteria in B2.
  Verify uniqueness: result count should be 2-4 (>=2 intermediates required - with only 1, the visit order is trivial).
  These become intermediate waypoints.

[B4. Compute Unconstrained-Optimal First (MANDATORY)]:
  **YOU MUST CALL THIS TOOL - DO NOT SKIP**
  Call `multi_stop_routing(stop_ids=[START_ID, ...intermediate IDs..., END_ID], start_id={START_ID}, end_id={END_ID}, ...)`.
  This returns ALL orderings with fixed start/end, sorted by total time/distance.
  **Record the #1 (unconstrained-optimal) ordering.**

[B5. Design CONFLICTING Constraints]:
  **CRITICAL**: Design constraint(s) that ELIMINATE the unconstrained-optimal ordering from B4.
  Use place data (opentime_week, logical dependencies, social insights) to justify constraints that force a different order.
  
  **VALIDATION**: Verify the unconstrained-optimal ordering VIOLATES your constraint. If it doesn't, redesign.

[B6. Select Constrained-Optimal]:
  From the sorted permutations in B4, find the FIRST ordering that satisfies ALL constraints.
  **FINAL CHECK**: It MUST differ from the unconstrained-optimal. If same, go back to B5.
  Extract the ACTUAL `total_distance_m` and `total_time_s`.

[B7. Question Formulation]:
  Construct a persona/scenario question with implicit criteria and constraints.
  Do NOT reveal that the constraint forces a detour.

[B8. Reference Verification (MANDATORY for social insights)]:
  **IF using social insights (long_mem/short_mem), YOU MUST CALL `search_notes`**
  Verify social insights with `search_notes`.

[B9. Final Output]:
  Include `waypoints` with `reason` and `ref`.
  **CRITICAL**: Use the EXACT values from tool response for total_distance_m and total_time_s.


---

# Examples

## Example 1 

skipped, see Figure: Example for Task Instance

## Example 2 

skipped, see Figure: Example for Task Instance

## Example 3 

skipped, see Figure: Example for Task Instance


---

**System has automatically sampled 10 places to help you get started:**

`sample_place(num=10)` returned:
```json
{
  "pois": [
    {
      "osm_id": "B001B1H2DS",
      "name": "Luoying Lake",
      "place_type": "Scenery Spot|Tourist Attraction",
      "address": "No. 16 Luojia Mountain",
      "opentime_week": "Monday-Sunday 00:00-24:00",
      "rating": 4.0,
      "cost": null,
      "tag": "Facility: Fountain; Rule: Shuttle Bus Stop",
      "alias": "Jian Lake",
      "long_mem": "Atmosphere: Quiet, mirror-like surface; Scenery: Water lilies, lotus, winter snow; Feedback: Beautiful view",
      "short_mem": "Crowd: Extremely crowded during Cherry Blossom season"
    },
    {
      "osm_id": "B0FFI211Q3",
      "name": "COACH (Chicony Plaza)",
      "place_type": "Shopping|Brand Bags and Suitcases Store",
      "address": "1F Chicony Plaza, No. 6 Luoyu Road (Near Metro Jiedaokou Station Exit B)",
      "opentime_week": "Monday-Sunday 10:00-22:00",
      "rating": 4.2,
      "cost": null,
      "tag": "Brand: COACH; Category: Luxury Bags; Facility: Mall Parking",
      "alias": "",
      "long_mem": "Atmosphere: Premium, clean; Service: Professional staff; Target: High-end shopping",
      "short_mem": "Event: Presale (2nd-23rd) up to 45% off, Official sale (24th-25th) spend 500 get 500; Special Hours: Open until 23:00 on 24th-25th"
    },
    {
      "osm_id": "B0LAJRQMJH",
      "name": "Sanqi Homebar (Jiedaokou)",
      "place_type": "Sports & Recreation|Pub",
      "address": "Room 2103, Tower C, Fuhua Building (90m from Metro Jiedaokou Station Exit B)",
      "opentime_week": "Tuesday-Sunday 19:00-02:00",
      "rating": 3.9,
      "cost": 138,
      "tag": "Drinks: Cocktails, Beer; Service: Karaoke, Board games; Facility: Private rooms, Terrace, Smoking area; Rule: Women-friendly, Reservation required, Closed on Mondays",
      "alias": "",
      "long_mem": "Atmosphere: Cozy, blues lighting, relaxing; Audience: Gen Z, university students, introvert-friendly; Service: Bartender recommendations; Feedback: Great cocktails, affordable",
      "short_mem": "Event: Valentine's Day limited edition cocktails available"
    },
    ... 7 samples skipped
  ]
}
```

Now, please generate a new data sample based on the sampled places above.
\end{tcblisting}
\captionof{figure}{The specialized system prompt utilized by the Data Engine Agent to orchestrate data retrieval and query generation.}
\label{box:data_agent_prompt}

\twocolumn

\end{document}